\documentclass[10pt,twocolumn,letterpaper]{article}

\usepackage{cvpr}              

\usepackage{booktabs,tabularx,xcolor,amssymb}
\usepackage{pifont}
\newcommand{\cmark}{\ding{51}}  
\newcommand{\xmark}{\ding{55}}  
\def \eg {\textit{e.g.}}
\def \ie {\textit{i.e.}}

\usepackage{graphicx}
\usepackage[table]{xcolor}  
\usepackage{colortbl}       
\usepackage{multirow}
\usepackage{capt-of}
\usepackage{listings}
\usepackage{xcolor}
\newcounter{magicrownumbers}

\newcommand \footnoteONLYtext[1]
{
    \let \mybackup \thefootnote
    \let \thefootnote \relax
    \footnotetext{#1}
    \let \thefootnote \mybackup
    \let \mybackup \imareallyundefinedcommand
}

\definecolor{cvprblue}{rgb}{0.21,0.49,0.74}
\usepackage[pagebackref,breaklinks,colorlinks,allcolors=cvprblue]{hyperref}

\def\paperID{4404} 
\def\confName{CVPR}
\def\confYear{2026}

\def \eg {\textit{e.g.}}
\def \ie {\textit{i.e.}}

\title{\raisebox{-0.3\height}{\includegraphics[width=0.6cm]{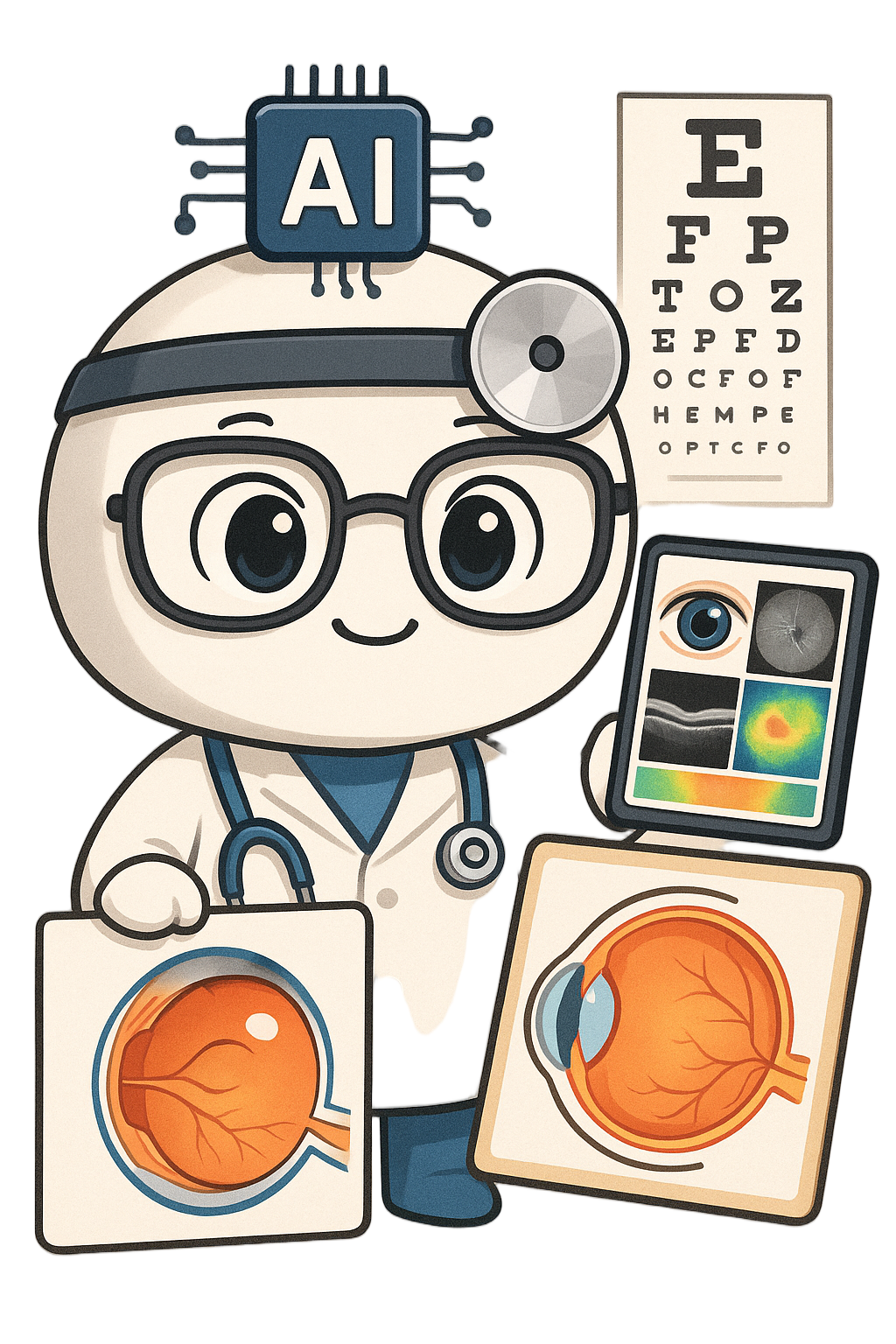}}
X-PCR: A Benchmark for Cross-modality Progressive Clinical Reasoning in Ophthalmic Diagnosis}

\author{
Gui Wang$^{1,3}$,~
Zehao Zhong$^{1}$,~
YongSong Zhou$^{1}$,~
Yudong Li$^{2}$,~
Ende Wu$^{5}$,~
Wooi Ping Cheah$^{3}$,~\\
Rong Qu$^{3}$,~
Jianfeng Ren$^{*,3}$,~
Linlin Shen$^{*,4,6}$~\\
\textsuperscript{\rm1}School of Computer Science and Software Engineering, Shenzhen University \quad \textsuperscript{\rm2}Tsinghua University \\ \textsuperscript{\rm3}University of Nottingham \quad
\textsuperscript{\rm4}School of AI, Shenzhen University \quad \textsuperscript{\rm5}Wenzhou Medical University \\
\textsuperscript{\rm6}Guangdong Provincial Key Laboratory of Intelligent Information Processing, Shenzhen University
 \\
{\tt\small jianfeng.ren@nottingham.edu.cn;~llshen@szu.edu.cn}
}

\begin{document}
\twocolumn[{
\renewcommand\twocolumn[1][]{#1}
\maketitle
\begin{center}
    \centering
    \captionsetup{type=figure}
    \includegraphics[width=0.98\linewidth]{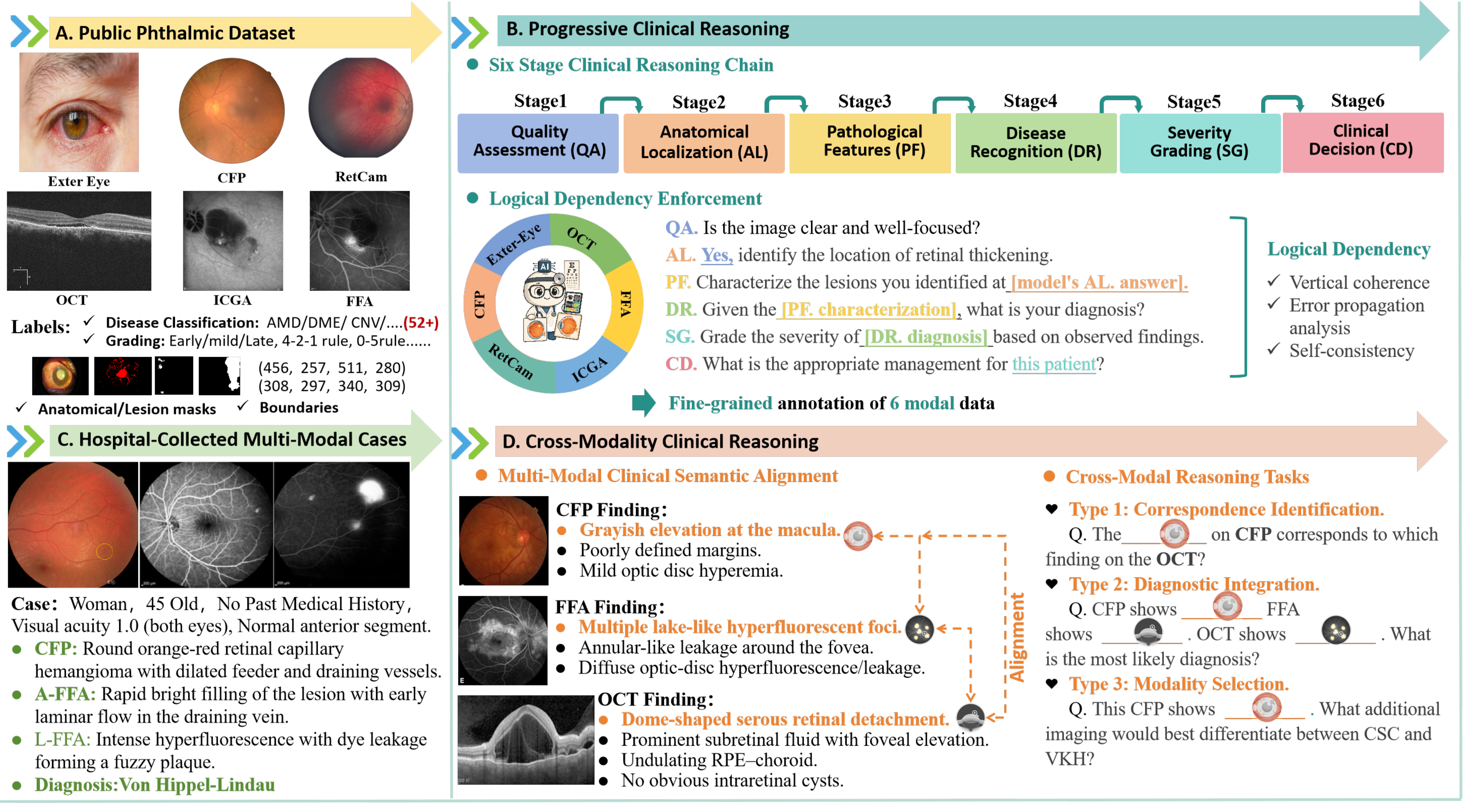}
    \caption{The X-PCR benchmark advances expert-level ophthalmic diagnosis with MLLMs through a \textbf{Six-Stage Progressive Clinical Reasoning Chain} (Block B) and a \textbf{Cross-Modal Clinical Reasoning} framework with precise modality alignment (Block D). Built on a foundation of \textbf{26,415} images across six modalities curated from 51 datasets (Block A), the benchmark integrates these into \textbf{multi-modal aligned} patient cases enriched with clinical context (Block C), modeling 52 diseases and yielding \textbf{177,868} expert-verified QA pairs.
    } 
    \label{overview}
\end{center}
}]

\footnoteONLYtext{*Corresponding author}
\begin{abstract}
Despite significant progress in Multi-modal Large Language Models (MLLMs), their clinical reasoning capacity for multi-modal diagnosis remains largely unexamined. Current benchmarks, mostly single-modality data, can't evaluate progressive reasoning and cross-modal integration essential for clinical practice. We introduce the Cross-Modality Progressive Clinical Reasoning (\textbf{X-PCR}) benchmark, the first comprehensive evaluation of MLLMs through a complete ophthalmology diagnostic workflow, with two reasoning tasks: 1) a six-stage progressive reasoning chain spanning image quality assessment to clinical decision-making, and 2) a cross-modality reasoning task integrating six imaging modalities. The benchmark comprises 26,415 images and 177,868 expert-verified VQA pairs curated from 51 public datasets, covering 52 ophthalmic diseases. Evaluation of 21 MLLMs reveals critical gaps in progressive reasoning and cross-modal integration. Dataset and code: https://github.com/CVI-SZU/X-PCR.  

\end{abstract}

\section{Introduction}
\label{sec:intro}

The advent of Multi-modal Large Language Models (MLLMs) has markedly accelerated the integration of artificial intelligence into clinical practice~\cite{moor2023foundation, acosta2022multimodal}. These models have demonstrated proficiency in interpreting medical images and generating diagnostic reports~\cite{tanno2025flamingoCXR,li2025braingpt}. Recent advances in large-scale multi-modal learning have enabled systems such as GPT-5~\cite{openai2025gpt5}, Gemini-2.5-pro~\cite{google2024gemini}, and several domain-specific medical models~\cite{liu2025llava, moor2025med-flamingo} to achieve remarkable performance on medical imaging tasks. However, a critical question persists: \textit{can current MLLMs orchestrate the systematic, multi-step, and multi-modal reasoning essential for real-world diagnostic decision-making?}

Visual question answering (VQA) has emerged as a prominent framework for evaluating MLLMs in medical diagnostics, spanning applications in pathology (PathVQA~\cite{he2020pathvqa}, WSI-VQA~\cite{WSIvqa}), surgery (EndoVis-VQLA~\cite{EndoVis-VQLA}, EndoBench~\cite{EndoBench}), and general medical domains (PMC-VQA~\cite{zhang2023pmc}, SLAKE~\cite{Slake}). However, these models are predominantly limited to single-modality, single-turn tasks. Although newer benchmarks such as MedFrameQA~\cite{yu2025medframeqa} and Rjua-Meddqa~\cite{jin2024rjua} have started incorporating progressive reasoning tasks like multi-image aggregation, they still lack a systematic evaluation of \textbf{multi-stage clinical reasoning} and \textbf{cross-modal integration}. 

This gap is particularly critical in ophthalmology, where diagnosing conditions like diabetic macular edema (DME) requires synthesizing multimodal findings, \eg, correlating microaneurysms on CFP with leakage on FFA and retinal structural changes on optical coherence tomography (OCT)~\cite{jin2025VLMoph,hajNajeeb2025dme}. Despite this clinical reality, most ophthalmic benchmarks evaluate imaging modalities in isolation and fail to capture the \textbf{cross-modal integrative reasoning} essential for diagnosis~\cite{wang2024multimodalAI,luo2025multimodaloph,jin2025VLMoph}. Furthermore, they prioritize diagnostic accuracy over predictive confidence, undermining reliable assessment of model safety and trustworthiness~\cite{kurz2022uncertainty,zhang2023calibration,lamberta2024trustworthy,goisauf2025trust}.

\begin{table*}[t]
\caption{Comparison of medical MLLM benchmarks. X-PCR provides the first framework for evaluating cross-modal progressive clinical reasoning across six key ophthalmic modalities, with uncertainty-aware assessment and expertise-graded evaluation.}
\label{tab:bench-compact}
\centering
\small
\setlength{\tabcolsep}{9pt}
\begin{tabularx}{\textwidth}{lrrrrccccccc}
\toprule
\textbf{Benchmark} & \textbf{Domain}  & \textbf{\#Modalities} & \textbf{\#Images} & \textbf{\#VQAs} & \textbf{PCR} & \textbf{MID} & \textbf{UA} & \textbf{EGE} \\
\midrule
PathVQA~\cite{he2020pathvqa}$_{\text{(CVPR'21)}}$ & Pathology & single & 5.0K & 32.8K & \xmark & \xmark & \xmark & \xmark\\
WSI-VQA~\cite{WSIvqa}$_{\text{(ECCV'24)}}$ & Pathology & single & 0.98K & 8.7K & \xmark & \xmark & \xmark & \xmark\\
PathBench~\cite{pathbench}$_{\text{(TMI'25)}}$ & Pathology & single & 5.3K & 6.3K & \xmark & \xmark & \xmark & \xmark\\
EndoVis-VQLA~\cite{EndoVis-VQLA}$_{\text{(ICRA'23)}}$ & Endoscopy & single & 1.5K & 9.5K & \xmark & \xmark & \xmark & \xmark\\
EndoBench~\cite{EndoBench}$_{\text{(NeurIPS'25)}}$ & Endoscopy & single & 6.8K & 6.8K & \xmark & \xmark & \xmark & \xmark\\
GEMeX~\cite{liu2025gemex}$_{\text{(ICCV'25)}}$ & Chest & single & 151K & 1605K & \xmark & \xmark & \xmark & \xmark\\
LOMD~\cite{lmod}$_{\text{(NAACL'25)}}$ & Ophthalmic & 5 & 21K & 21K & \xmark & \xmark & \xmark & \xmark\\
SLAKE~\cite{lau2018dataset}$_{\text{(ISBI'21)}}$ & General & 3 & 643 & 14K &\xmark & \xmark & \xmark & \xmark\\
OmniMedVQA~\cite{Omnimedvqa}$_{\text{(CVPR'24)}}$ & General & 12 & 118K & 128K &\xmark & \xmark & \xmark & \xmark\\
PMC-VQA~\cite{zhang2024development}$_{\text{(NatureCM'24)}}$ & General & 5 & 149K & 227K & \xmark & \xmark & \xmark & \xmark\\
\midrule
Medical-CXR-VQA~\cite{hu2024interpretable}$_{\text{(MIA'24)}}$ & Chest & single & 227K & 377K & \cmark & \xmark & \xmark & \xmark\\
MedThink~\cite{gai2024medthink}$_{\text{(ArXiv'24)}}$ & General & single & 2.8K & 9.2K & \cmark & \xmark & \xmark & \xmark\\
Rjua-Meddqa \cite{jin2024rjua}$_{\text{(ArXiv'24)}}$ & General & single & 2K & 72K & \cmark & \xmark & \xmark & \xmark\\
MedFrameQA~\cite{yu2025medframeqa}$_{\text{(ArXiv'25)}}$ & General & single & 2.8K & 9.2K & \cmark & \xmark & \xmark & \xmark\\
MicroVQA~\cite{microvqa}$_{\text{(CVPR'25)}}$ & Microscopy & 3 & 423 & 1.04K & \cmark & \xmark & \xmark & \xmark\\
EyePCR~\cite{eyepcr}$_{\text{(ArXiv'25)}}$ & Ophthalmic & single & 2.1K & 210K & \cmark & \xmark & \xmark & \xmark\\
\rowcolor{gray!12}
\textbf{X-PCR$_{\text{(Ours)}}$} & \textbf{Ophthalmic} & \textbf{6} & \textbf{26.4k} & \textbf{177.9k} & \cmark & \cmark & \cmark & \cmark \\
\bottomrule
\end{tabularx}
\vspace{2pt}
\footnotesize
\textbf{PCR}: Progressive Clinical Reasoning. \textbf{MID}: Multimodal Integrate Diagnostic. \textbf{UA}: Uncertainty-Aware; \textbf{EGE}: Expertise-Graded Evaluation.
\end{table*}

To bridge these critical gaps, we introduce the Cross-Modality Progressive Clinical Reasoning (\textbf{X-PCR}) benchmark, a comprehensive framework designed to evaluate MLLMs across the complete diagnostic reasoning spectrum in ophthalmology, as illustrated in Fig.~\ref{overview}. X-PCR systematically addresses the fundamental challenges in clinical MLLMs evaluation through an integrated approach.

First, to address the challenge that real-world diagnosis requires logically-sequenced multi-step reasoning rather than isolated task performance, X-PCR formalizes clinical diagnosis as a six-stage reasoning chain, from image quality assessment to anatomical localization, lesion characterization, disease diagnosis, severity grading, and eventually clinical decision-making, which mirrors the systematic workflow of ophthalmologists. Unlike existing benchmarks~\cite{pathbench, WSIvqa, EndoBench}, our framework enforces causal dependencies between stages, \eg, requiring models to correctly localize lesions before characterizing them and establish diagnoses before determining severity. This approach thus evaluates diagnostic validity and reasoning coherence, advancing from isolated metrics to progressive clinical reasoning that reflects genuine decision-making. 

Second, recognizing that ophthalmologic diagnoses often depend on synthesizing evidence from multiple imaging modalities, a capability neglected by current benchmarks~\cite{lmod, eyepcr}, X-PCR integrates six key ophthalmic modalities, \ie, \textit{External Photography} (\textit{EP}), \textit{Color Fundus Photography} (\textit{CFP}), \textit{Fluorescein Angiography} (\textit{FFA}), \textit{Indocyanine Green Angiography} (\textit{ICGA}), \textit{Optical Coherence Tomography} (\textit{OCT}) and \textit{RetCam}, with explicitly annotated cross-modal correspondences. For each disease presentation, we provide ground-truth linkages, such as associating hemorrhages on CFP with specific OCT findings, enabling direct evaluation of cross-modal reasoning. This design allows us to assess not only single-modality interpretation but also a model’s ability to integrate multi-modal evidence, a critical skill for accurate and context-aware diagnosis.

To support cross-modal progressive clinical reasoning, we construct the X-PCR benchmark comprising 26,415 expert-verified images and 177,868 VQA pairs from 51 ophthalmic datasets, covering 52 diseases, and enhance clinical realism with 58 multi-modal aligned patient cases from partner hospitals. Unlike conventional benchmarks using overall accuracy, X-PCR introduces a difficulty- and uncertainty-aware evaluation scheme, stratifying questions into resident, attending, and specialist tiers and incorporating self-reported confidence to compute an \textit{Uncertainty-Aware Score} (\textit{UAS}) and \textit{Expected Calibration Error} (\textit{ECE}), thereby quantifying both accuracy and reliability while penalizing overconfident errors in high-stakes scenarios.

Our main contributions are four-fold: 
1)~We introduce the X-PCR benchmark, the first comprehensive framework for evaluating diagnostic reasoning of MLLMs in ophthalmology, comprising 26,415 expert-verified images and 177,868 VQA pairs spanning 52 diseases to enable rigorous assessment of clinical reasoning capabilities. 
2)~X-PCR formalizes diagnosis as a six-stage reasoning chain with enforced causal dependencies, enabling assessment of diagnostic validity and reasoning coherence beyond isolated tasks. 
3)~X-PCR integrates six ophthalmic imaging modalities with explicit ground-truth correspondences, supporting systematic evaluation of multimodal clinical reasoning.
4)~Evaluation of 21 leading MLLMs on X-PCR reveals a substantial performance gap in progressive reasoning and cross-modal diagnosis compared to clinical expertise. 

\section{Related Work}
\label{sec:formatting}

\subsection{Multimodal Large Language Models}  
\textbf{General-Purpose MLLMs.} \quad
Recent advances in large scale vision language pretraining have yielded models with strong zero-shot capabilities. Systems such as LLaVA \cite{liu2025llava}, BLIP-3 \cite{li2025blip3}, and Flamingo \cite{alayrac2022flamingo} combine frozen large language models with visual encoders to achieve state-of-the-art results, while newer models including GPT-5 \cite{openai2025gpt5} and LLaMA-4 \cite{meta2025llama4} demonstrate emergent medical image understanding even without domain specific fine-tuning. Nevertheless, evaluations on general-purpose VQA benchmarks \cite{goyal2017making,hudson2019gqa} fail to assess the progressive, clinically grounded reasoning for real-world diagnostic decision making.


\noindent\textbf{Medical-Specific MLLMs.} \quad
Medical MLLMs address the domain gaps via continued pretraining and instruction tuning on clinical data. Representative models include LLaVA-Med~\cite{li2025llava-med}, Med-Flamingo~\cite{moor2025med-flamingo}, Med-PaLM M~\cite{tu2025medpalm-m}, HuaTuoGPT~\cite{chen2024huatuogpt}, Lingshu~\cite{xu2025lingshu}, and MedGemma~\cite{medgemma}, which adapt general-purpose models using biomedical corpora, modality-specific encoders, or multimodal medical data. Although medical MLLMs exhibit enhanced knowledge, their evaluation remains confined to classification accuracy, neglecting multi-step reasoning. Our analysis confirms that pretraining improves knowledge but fails to instill systematic diagnostic reasoning.


\noindent\textbf{Ophthalmic MLLMs.} \quad
Ophthalmology has emerged as a leading domain for medical MLLM development. RETFound~\cite{zhou2023foundation} establishes a retinal foundation model pretrained on 1.6 million images, demonstrating strong transfer learning across ophthalmic tasks. Subsequent systems like ChatPGT~\cite{madadi2025chatgpt}, Ophtha-LLaMA2~\cite{zhao2023ophtha}, and OphGLM~\cite{deng2024ophglm} have advanced toward ophthalmology-specific vision language assistants through instruction tuning, supporting tasks such as case based dialogue and image-guided QA. However, these models remain limited to single-modality inputs and isolated classification tasks~\cite{li2025visionunite, zhou2023foundation}. Our benchmark directly addresses the critical gap between such single-task systems and the multi-modality, chain-like reasoning required for clinical diagnostic workflows. 


\subsection{Benchmarks for Medical-Specialized MLLMs}
Medical VQA has emerged as the primary evaluation paradigm for medical MLLMs, with specialized benchmarks developed across clinical domains. As shown in Table~\ref{tab:bench-compact}, PathVQA~\cite{he2020pathvqa}, WSI-VQA~\cite{WSIvqa}, and PathBench~\cite{pathbench} focus on pathology with large-scale open-ended QA on microscopy images; EndoVis-VQLA~\cite{EndoVis-VQLA} and EndoBench~\cite{EndoBench} target endoscopic scenarios with procedure centric queries; GEMeX~\cite{liu2025gemex} provides a large scale chest X-ray benchmark with 1.605M QA pairs. General-domain benchmarks like SLAKE~\cite{Slake}, OmniMedVQA~\cite{Omnimedvqa}, and PMC-VQA~\cite{zhang2023pmc} offer broader coverage, while LOMD~\cite{lmod} specifically addresses ophthalmology. Despite some benchmarks incorporating multi-modal annotations, evaluation remains predominantly single-modality and single-step, with no systematic assessment of multi-step clinical reasoning. 


Recent efforts focus on clinical reasoning assessment. Medical-CXR-VQA~\cite{hu2024interpretable} establishes a chest X-ray dataset with LLM-assisted QA generation and expert validation, evaluating abnormality, location, and lesion recognition. 
Rjua-Meddqa~\cite{jin2024rjua} focuses on multimodal medical document QA, targeting layout understanding, numerical reasoning, and clinical reasoning. 
MedFrameQA~\cite{yu2025medframeqa} evaluates multi-image reasoning using 2–5 image question pairs, measuring performance degradation when evidence requires cross-frame aggregation. 
EyePCR~\cite{eyepcr} assesses ophthalmic surgical cognition across perception, comprehension, and reasoning stages using 210K VQA pairs and knowledge-grounded tasks. 
Unlike existing works, X-PCR structures reasoning as a dependent chain where later stages require correct prior outputs, evaluating reasoning coherence, cross-modal integration, and calibration.



\section{X-PCR Benchmark}
We introduce the \textbf{X-PCR} (Cross-modality Progressive Clinical Reasoning) benchmark, a comprehensive framework for evaluating MLLMs across the full diagnostic reasoning spectrum in ophthalmology. X-PCR addresses three core challenges: 1) modeling progressive reasoning with explicit logical dependencies; 2) enabling cross-modality clinical alignment; and 3) providing difficulty stratified and uncertainty-aware evaluation. 

\subsection{Progressive Clinical Reasoning} 
\noindent \textbf{Six-Stage Clinical Reasoning Chain.} \quad 
We formalize ophthalmic diagnosis as a six-stage progressive reasoning chain, where each stage builds logically upon its predecessor, establishing causal dependencies throughout the diagnostic workflow. 
\textbf{1)~Image Quality Assessment} (\textbf{IQA}), assessing whether image quality meets diagnostic standards by identifying technical flaws (\eg, poor focus, illumination issues, artifacts) that compromise reliability, and determining if an image supports confident diagnosis or requires reacquisition. IQA gates the entire pipeline: non-diagnostic images are reacquired or down-weighted, while only adequate images proceed next to prevent error propagation. 
\textbf{2)~Anatomical Localization} (\textbf{AL}), requiring precise identification of key fundus structures (optic disc, macula, vascular arcades, peripheral retina) to establish anatomical context for pathology interpretation. Lesion Characterization depends entirely on the anatomical map, requiring spatial anchoring of lesion descriptors to AL-defined regions, with AL inaccuracies propagating as uncertainty in subsequent stages. 
\textbf{3)~Lesion Characterization} (\textbf{LC}), characterizing lesion morphology and distribution using precise clinical terminology (\eg, ``flame-shaped hemorrhages," ``cotton-wool spots"), and providing foundational evidence for diagnosis. Disease Diagnosis depends entirely on spatially anchored LC findings; incomplete or contradictory LC data broadens the differential diagnosis or reduces diagnostic confidence. 
\textbf{4)~Disease Diagnosis} (\textbf{DD}), evaluating diagnostic synthesis by integrating prior observations into a ranked differential diagnosis with evidence-based justification. Severity Grading depends entirely on the DD outcome; any diagnostic change triggers re-evaluation using appropriate clinical rubrics, with SG conclusions required to align with DD-implied pathological burden. 
\textbf{5)~Severity Grading} (\textbf{SG}).  
Following diagnosis, models must apply disease-specific severity scales (\eg, International Clinical Diabetic Retinopathy (ICDR)  Scale) using clinical criteria to quantify progression. Clinical decisions map directly from DD and SG outcomes; management intensity scales with severity, and low confidence triggers conservative actions like confirmatory testing. 
\textbf{6)~Clinical Decision-Making} (\textbf{CD}), assessing the translation of diagnostic conclusions into actionable clinical management, including treatment selection, referral urgency, and follow-up planning, supported by clinical rationale. CD integrates outputs from all prior stages, deriving decisions directly from DD and SG to ensure end-to-end clinical workflow coherence.

\noindent \textbf{Logical Dependency of Reasoning Chain.} \quad 
Unlike prior benchmarks evaluating stages in isolation, X-PCR enforces explicit logical dependencies across the diagnostic workflow. As illustrated in Fig.~\ref{overview}(B), each stage conditions its reasoning on validated outputs from the preceding step, \eg, \textit{\textbf{IQA}} assesses image quality; \textit{\textbf{AL}} localizes anatomy conditional on \textit{\textbf{IQA}}; \textit{\textbf{LC}} characterizes lesions based on \textit{\textbf{AL}}; \textit{\textbf{DD}} synthesizes findings from \textit{\textbf{LC}}; \textit{\textbf{SG}} grades severity given \textit{\textbf{DD}}; and \textit{\textbf{CD}} derives management from \textit{\textbf{SG}}. 

This dependency-aware design enables evaluation of three critical reasoning properties: 1)~\textit{vertical coherence}, whether downstream predictions logically follow upstream conclusions; 2)~\textit{error propagation}, the extent to which upstream errors degrade downstream performance; and 3)~\textit{self-consistency}, the stability of a model's diagnostic narrative across stages. 
The framework shifts evaluation from isolated accuracy to end-to-end reasoning integrity, assessed by both stage-wise accuracy and chain completion rate.

\begin{figure*}[t]
\centering
\includegraphics[width=1\linewidth]{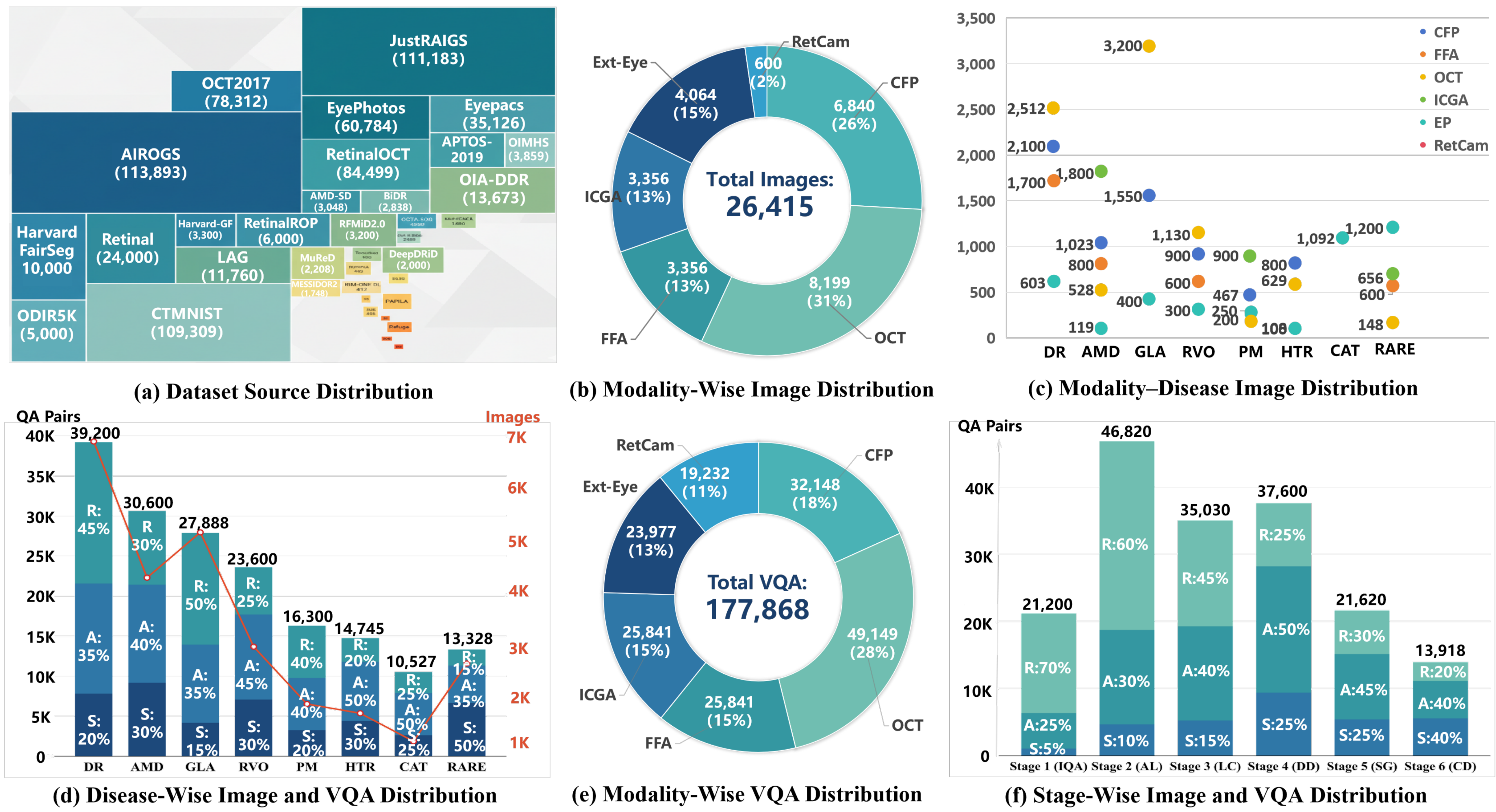}
\caption{
X-PCR: 26,415 images and 177,868 VQAs from 51 datasets (as in (a)), spanning 6 modalities (as in (b) and (e)), 8 disease categories (as in (c) and (d)), and 6 reasoning stages (as in (f)). It establishes a disease-centric benchmark for multimodal clinical alignment, balancing common and rare conditions according to real-world prevalence. Stratified by clinical difficulty and organized into diagnostic-aligned reasoning stages, it enables systematic evaluation of progressive, multimodal reasoning beyond isolated accuracy.
}
\label{VQA-Sta}
\end{figure*}


\subsection{Cross-Modality Clinical Reasoning} 
\noindent \textbf{Multi-Modal Semantic Alignment.} \quad
Ophthalmic diagnosis requires integrating complementary imaging modalities, each capturing distinct anatomical and pathophysiological characteristics: \textit{External Photography} (\textit{EP}) documents anterior segment anatomy, eyelid pathology, and ocular alignment; \textit{Color Fundus Photography} (\textit{CFP}) visualizes optic disc, retinal vasculature, and intraretinal lesions; \textit{Fluorescein Angiography} (\textit{FFA}) assesses vascular perfusion and leakage; \textit{Indocyanine Green Angiography} (\textit{ICGA}) delineates choroidal circulation and polypoidal lesions; \textit{Optical Coherence Tomography} (\textit{OCT}) reveals cross-sectional retinal microstructure; and \textit{RetCam} Imaging wide-field visualization of the peripheral retina in pediatric and neonatal populations. Together, these modalities provide a comprehensive, multi-scale view of ocular structure and function.


To address fragmented and unaligned evidence in current ophthalmic multi-modal datasets (\eg, lacking semantic alignment to disease instances and temporal synchronization across eyes), we introduce a disease-centric framework with two key designs: 
1)~\textbf{Standardized Clinical Representation}: Each disease is defined by a unified \textit{Disease–Anatomy–Lesion–Grading–Management} schema and a modality$\times$evidence matrix specifying cross-modal anchors and terminology mappings; 
and 2)~\textbf{Temporally Aligned Multi-Modal Integration}: Contemporaneous same-eye images are structurally and semantically aligned into standardized exemplars, validated by a six-stage QA process ensuring narrative coherence and grading consistency across modalities for end-to-end clinical reasoning. 
For example, in \textit{diabetic macular edema}, aligned \textit{CFP}, \textit{FFA}, and \textit{OCT} inputs support a staged reasoning chain from lesion confirmation (Stage 2/3) to disease diagnosis (Stage 4), severity grading (Stage 5), and anti-VEGF treatment (Stage 6), ensuring semantically consistent and temporally coherent diagnostic evidence for reliable clinical reasoning.

\noindent \textbf{Cross-Modality Reasoning.} \quad 
We operationalize cross-modal reasoning through three progressively complex task families, forming a progressive pipeline from single-view matching to multi-view synthesis and decision support: 
1) \textbf{Correspondence Identification}, linking semantically equivalent findings across paired modalities (\eg, \textit{CFP} macular elevation $\leftrightarrow$ \textit{OCT} SRD) to establish spatial-semantic alignments; 
2) \textbf{Diagnostic Integration}, synthesizing complementary cues from aligned multi-modal inputs (\eg, \textit{CFP}, \textit{FFA} and \textit{OCT} in Fig.~\ref{overview}(D)) to infer the most probable diagnosis; 
and 3) \textbf{Modality Selection}, recommending the next informative imaging modality when diagnostic uncertainty persists (\eg, \textit{ICGA} to differentiate \textit{CSC}/\textit{VKH}), reflecting value-of-information reasoning in clinical workflows. These tasks form a pipeline where Correspondence supplies aligned evidence for Integration, and Modality Selection is triggered by residual uncertainty, enabling a progression from cross-modal matching to evidence synthesis to actionable decision support.

\subsection{Dataset Collection and VQA Generation}
\noindent \textbf{Public Dataset Curation.} \quad
We curate \textbf{X-PCR} from 51 public ophthalmic datasets covering 8 disease categories: \textit{diabetic retinopathy} (\textit{DR}), \textit{glaucoma} (\textit{GLA}),  \textit{cataract} (\textit{CAT}),  \textit{age-related macular degeneration} (\textit{AMD}),  \textit{hypertensive retinopathy} (\textit{HTR}),  \textit{pathological myopia} (\textit{PM}),  \textit{Retinal Vein Occlusion} (\textit{RVO}), and  \textit{rare conditions} (\textit{Rare}).
A unified quality control pipeline was applied: images with original labels and acceptable quality formed the VQA set, while low-resolution or severely artifacted images were reserved as adversarial samples without quality labels. To mitigate sampling bias, the dataset ensures representativeness across disease categories, severity stages (early/mid/late), and six imaging modalities. 


\noindent \textbf{Hospital-Collected Multi-Modal Cases.} \quad 
To ensure clinical realism, we prospectively collected 58 multimodal cases from ophthalmic hospitals. 
Eligible cases met four criteria: complete multi-modal workup, definitive diagnosis confirmed clinically or surgically, documented history and examination findings, and IRB approval with informed consent for research use. Each case includes a structured clinical vignette (\eg, age, symptoms, visual acuity, and relevant medical history), synchronized multi-modal imaging series across the six complementary modalities, and expert annotations covering cross-modal correspondences, differential diagnoses, and treatment recommendations. All cases were validated by attending ophthalmologists.


\noindent \textbf{VQA Generation.} \quad
We use a semi-automated pipeline that combines prompt-based generation with expert validation to unify schemas across 51 datasets for lesion segmentation, classification, and disease grading. 
Designed with ophthalmologists, each stage includes 2–10 question templates forming multiple-choice QA pairs, where templates are derived from original category definitions, and distractors are sampled from same attribute nouns to ensure semantic relevance. 
In the first five stages, VQAs are automatically generated by GPT-5~\cite{openai2025gpt5} from dataset labels and lesion annotations, with Gemini-2.5-pro~\cite{google2024gemini} auditing all content for consistency and accuracy. In the last stage of Clinical Decision-Making, which requires fine-grained clinical reasoning, VQAs are manually authored by ophthalmologists. 
ophthalmologists then assign difficulty (R/A/S) and clinical importance scores to each question type. 
For quality control, 20\% of VQA items undergo independent review by a second ophthalmologist. Adjudication is conducted to reach a consensus for items with inter-rater agreement below the threshold ($\kappa<0.8$); persistently disputed items are excluded. 
This workflow ensures scalability while maintaining clinical validity and interpretability.


\noindent \textbf{Dataset Statistics.} \quad
Fig.~\ref{VQA-Sta} summarizes the dataset composition, comprising 26,415 images and 177,868 VQA pairs curated from diverse public sources. The dataset spans 1) a broad spectrum of ophthalmic diseases categorized into eight groups; 2) six imaging modalities, dominated by \textit{OCT} (31\%) and \textit{CFP} (26\%), which together contribute the majority of VQA instances; and 3) a six-stage clinical reasoning pathway, from image quality assessment to clinical decision-making, with difficulty levels (S/A/R) calibrated per disease and stage to ensure discriminative evaluation and scalability. 
This design emphasizes clinical relevance and cross-modal consistency in a multi-task framework. Expanded dataset details are provided in Appendix. 


\subsection{Difficulty- and Uncertainty-Aware Assessment}
\label{sec:diff-uncertainty}
\noindent \textbf{Difficulty-Aware Metrics.} \quad
We employ a three-tier difficulty framework aligned with clinical training: \textit{Resident} (\textit{R}), \textit{Attending} (\textit{A}), and \textit{Specialist} (\textit{S}) levels, corresponding to foundational recognition, contextual diagnosis, and subspecialty interpretation. Each case is assigned an importance score reflecting clinical impact, \eg, risk of missing sight threatening disease. 
Then, we define the following evaluation metrics: 1) \textit{Stage-Wise Accuracy} (\textit{SWA}) per reasoning stage; 2) \textit{Chain Completion Rate} (\textit{CCR}) for fully correct six-stage chains; 3) \textit{Expertise-Stratified Accuracy} (\textit{ESA}) within each difficulty tier. 



\noindent \textbf{Uncertainty-Aware Metrics.} \quad
Model responses include self-reported confidence (normalized to $[0,1]$), categorized as \textit{Correct Confident} (\textit{CC}), \textit{Correct Uncertain} (\textit{CU}), \textit{Incorrect Uncertain} (\textit{IU}), or \textit{Incorrect Confident} (\textit{IC}). Each category receives a base score, aggregated with difficulty–impact weights into a weighted \textit{Uncertainty-Aware Score} (\textit{UAS}), rewarding calibrated confidence and penalizing overconfident errors.
We further report \textit{Expected Calibration Error} (\textit{ECE}), measuring bin-wise discrepancies between confidence and empirical accuracy. Together, \textit{UAS} and \textit{ECE} evaluate both prediction quality and uncertainty calibration in clinically stratified settings.



\begin{table*}[t]
\centering
\caption{\textbf{Benchmarking MLLMs on X-PCR.} 
Models evaluated across stage-wise accuracy, chain completion, expertise stratification, and uncertainty aware metrics.
Human baselines shown in gray rows. Best results in \textbf{bold}, second-best \underline{underlined}. 
}
\label{tab:main_results}
\resizebox{\linewidth}{!}{
\begin{tabular}{clcccccccccccccc} 
\toprule
& \multirow{2}{*}{\textbf{Model}} &
\multicolumn{7}{c}{\textbf{Stage-Wise Acc. (\%) $\uparrow$}} &
\multicolumn{4}{c}{\textbf{Expertise-Stratified Acc. (\%)} $\uparrow$} &
\multirow{2}{*}{\textbf{CCR (\%)}} &
\multicolumn{2}{c}{\textbf{Uncertainty}} \\
\cmidrule(lr){3-9} \cmidrule(lr){10-13} \cmidrule(lr){15-16}
& & \textbf{IQA} & \textbf{AL} & \textbf{LC} & \textbf{DD} & \textbf{SG} & \textbf{CD} & \textbf{Avg}
& \textbf{R-level} & \textbf{A-level} & \textbf{S-level} & \textbf{Avg}
&  & \textbf{ECE} $\downarrow$ & \textbf{UAS} $\uparrow$ \\
\midrule
\multirow{6}{*}{\rotatebox[origin=c]{90}{\textbf{\textit{Commercial}}}} & 
GPT-5~\cite{openai2025gpt5}& \textbf{98.90} & \textbf{88.05} & \textbf{81.02} & \underline{73.11} & \textbf{61.65} & \textbf{54.71} & \textbf{76.24} & \textbf{80.64} & \textbf{76.19} & \textbf{62.92} & \textbf{73.25} & \textbf{24.47} & \textbf{0.062} & \textbf{74.32} \\
& Gemini-2.5-Pro ~\cite{google2024gemini} & \underline{94.58} & \underline{84.05} & \underline{71.15} & 72.48 & \underline{58.07} & 37.35 & \underline{69.61} & 75.71 & \underline{70.68} & \underline{54.76} & \underline{67.05} & 0.77 & \underline{0.072} & \underline{68.26} \\
& GLM-4.5v ~\cite{glm45v} & 90.56 & 81.74 & 65.32 & 70.75 & 50.47 & 46.58 & 67.57 & \underline{77.91} & 69.19 & 47.45 & 64.85 & 0.12 & 0.091 & 65.33 \\
& GPT-5-nano~\cite{openai_gpt5_nano_docs_2025}& 91.30 & 77.82 & 65.08 & 64.41 & 53.89 & \underline{48.92} & 66.90 & 73.23 & 69.28 & 50.75 & 64.42 & \underline{3.46} & 0.076 & 64.63 \\
& Gemini-2.5-Flash ~\cite{google2025gemini25flash} & 89.68 & 82.04 & 62.82 & \textbf{74.68} & 49.27 & 36.28 & 65.80 & 73.29 & 65.48 & 49.03 & 62.60 & 0.69 & 0.087 & 66.24 \\
& Claude-Haiku-4.5 ~\cite{haiku45_2025_news} & 77.59 & 75.90 & 62.79 & 63.23 & 47.73 & 43.83 & 61.84 & 65.50 & 59.89 & 50.48 & 58.62 & 0.12 & 0.086 & 60.91 \\
\midrule
\multirow{10}{*}{\rotatebox[origin=c]{90}{\textbf{\textit{Open-Source}}}} & 
Qwen2.5-VL-72B~\cite{Qwen2.5-VL} & \underline{93.13} & \underline{81.54} & 72.51 & \underline{67.15} & \underline{53.52} & \textbf{47.06} & \underline{69.15} & \underline{74.89} & 70.11 & \textbf{55.69} & \textbf{66.90} & \textbf{13.77} & \underline{0.075} & \textbf{66.37} \\
& Qwen3-VL-30B-A3B~\cite{xu2025qwen3}& 89.25 & 79.33 & \underline{72.89} & 65.14 & 51.51 & \underline{44.40} & 67.09 & 72.80 & \textbf{70.84} & 52.23 & 65.29 & \underline{1.04} & \textbf{0.074} & \underline{64.63} \\
& InternVL-32B ~\cite{internvl2024} & \textbf{94.35} & \textbf{83.25} & \textbf{74.17} & \textbf{71.01} & \textbf{54.82} & 43.22 & \textbf{70.14} & \textbf{77.10} & \underline{70.37} & \underline{52.37} & \underline{66.61} & 0.92 & 0.086 & 64.46 \\
& LLaVA-v1.5-13B ~\cite{liu2023llava15}& 59.64 & 58.20 & 52.37 & 44.14 & 34.29 & 39.37 & 48.00 & 55.17 & 51.02 & 35.75 & 47.31 & 0.06  & 0.102 & 45.78 \\
\cmidrule(lr){2-16}
& Qwen3-VL-8B~\cite{xu2025qwen3}& \textbf{89.40} & 64.67 & 75.80 & \underline{67.84} & \underline{49.09} & \textbf{41.49} & \underline{64.72} & \underline{72.73} & 61.49 & \textbf{48.93} & 61.05 & \underline{0.45} & 0.093 & \textbf{61.51} \\
& Qwen2.5-VL-7B ~\cite{Qwen2.5-VL} & \underline{88.34} & \underline{68.28} & \underline{78.15} & \textbf{72.04} & 48.14 & \underline{38.31} & \textbf{65.54} & \textbf{74.81} & \textbf{68.43} & 45.55 & \textbf{62.93} & 0.24 & \textbf{0.074} & 63.11 \\
& LLaMA3-LLaVA-Next-8B ~\cite{llava_next_interleave_2024} & 69.99 & 61.24 & \textbf{79.44} & 47.60 & 44.86 & 38.27 & 56.90 & 67.83 & 53.67 & 40.48 & 53.99 & 0.00 & 0.101 & 53.22 \\
& InternVL-8B ~\cite{internvl2024} & 87.57 & \textbf{73.97} & 69.42 & 65.53 & \textbf{53.91} & 37.22 & 64.60 & 70.73 & \underline{67.30} & \underline{48.33} & \underline{62.12} & \textbf{0.63} & \underline{0.083} & \underline{62.75} \\
& ShareGPT4V-7B ~\cite{chen2024sharegpt4v} & 65.93 & 63.03 & 53.70 & 46.12 & 34.62 & 34.87 & 49.71 & 67.51 & 50.09 & 29.41 & 49.00 & 0.00 & 0.101 & 48.39 \\
& LLaVA-v1.5-7B~\cite{liu2023llava15} & 60.40 & 63.30 & 56.53 & 45.42 & 29.31 & 33.29 & 48.04 & 66.71 & 49.10 & 24.90 & 46.90 & 0.00  & 0.102 & 45.64 \\
\midrule
\multirow{5}{*}{\rotatebox[origin=c]{90}{\textbf{\textit{Medical}}}} & 
MedGemma-27B-IT ~\cite{medgemma}& \textbf{90.03} & \textbf{72.44} & \underline{72.28} & \underline{64.27} & 46.94 & \textbf{45.32} & \textbf{65.21} & \textbf{74.19} & \underline{64.63} & \textbf{46.14} & \textbf{61.65} & 0.06 & \textbf{0.087} & \textbf{61.74} \\
& HuatuoGPT-Vision-34B ~\cite{chen2024huatuogpt} & 75.94 & 66.01 & 55.56 & 58.48 & \underline{47.59} & \underline{40.64} & 57.37 & 64.44 & 59.31 & \underline{41.95} & 55.23 & \textbf{0.21} & 0.103 & 53.88 \\
& Lingshu-7B Lingshu-7B~\cite{xu2025lingshu}& \underline{80.50} & \underline{66.20} & \textbf{73.37} & \textbf{67.41} & \textbf{49.64} & 32.81 & \underline{61.66} & \underline{70.18} & \textbf{65.15} & 41.89 & \underline{59.07} & \underline{0.18} & \underline{0.092} & \underline{58.18} \\
& LLaVA-Med-7B ~\cite{li2025llava-med} & 50.94 & 37.19 & 55.57 & 46.18 & 20.61 & 27.12 & 39.60 & 44.01 & 40.79 & 27.48 & 37.43 & 0.00 & 0.096 & 37.05 \\
& HuatuoGPT-Vision-7B ~\cite{chen2024huatuogpt}  & 61.95 & 56.42 & 48.36 & 59.21 & 44.61 & 36.75 & 51.22 & 54.22 & 55.23 & 39.69 & 49.71 & 0.06  & 0.103 & 48.99 \\
\midrule
\multirow{3}{*}{\rotatebox[origin=c]{90}{\textbf{\textit{Expert}}}}
& \cellcolor{gray!15} Residents (n=23)  & \cellcolor{gray!15} 90.77 & \cellcolor{gray!15} 80.68 & \cellcolor{gray!15} 69.82 & \cellcolor{gray!15} 63.73 & \cellcolor{gray!15} 53.91 & \cellcolor{gray!15} 47.22 & \cellcolor{gray!15} 67.69 & \cellcolor{gray!15} 82.43 & \cellcolor{gray!15} 71.34 & \cellcolor{gray!15} 48.17 & \cellcolor{gray!15} 67.31 & \cellcolor{gray!15} 8.39 & \cellcolor{gray!15} 0.128 & \cellcolor{gray!15} 63.78 \\
& \cellcolor{gray!15} Attendings (n=10) & \cellcolor{gray!15} \underline{95.46} & \cellcolor{gray!15} \underline{87.69} & \cellcolor{gray!15} \underline{81.51} & \cellcolor{gray!15} \underline{77.15} & \cellcolor{gray!15} \underline{70.59} & \cellcolor{gray!15} \underline{67.06} & \cellcolor{gray!15} \underline{79.91} & \cellcolor{gray!15} \underline{92.18} & \cellcolor{gray!15} \underline{78.76} & \cellcolor{gray!15} \underline{68.27} & \cellcolor{gray!15} \underline{79.74} & \cellcolor{gray!15} \underline{41.24} & \cellcolor{gray!15} \underline{0.091} & \cellcolor{gray!15} \underline{77.16} \\
& \cellcolor{gray!15} Specialists (n=8) & \cellcolor{gray!15} \textbf{97.80} & \cellcolor{gray!15} \textbf{90.05} & \cellcolor{gray!15} \textbf{85.32} & \cellcolor{gray!15} \textbf{80.11} & \cellcolor{gray!15} \textbf{72.85} & \cellcolor{gray!15} \textbf{70.97} & \cellcolor{gray!15} \textbf{82.85} & \cellcolor{gray!15} \textbf{95.88} & \cellcolor{gray!15} \textbf{91.28} & \cellcolor{gray!15} \textbf{83.63} & \cellcolor{gray!15} \textbf{90.26} & \cellcolor{gray!15} \textbf{62.48} & \cellcolor{gray!15} \textbf{0.063} & \cellcolor{gray!15} \textbf{90.63} \\
\bottomrule
\end{tabular}
}
\end{table*}

\section{Experimental Results}

\subsection{Experimental Setup}
\noindent \textbf{Evaluated Models.} \quad 
We evaluate 21 MLLMs, including 6 commercial, 10 open-source, and 5 med-specialized models across varying capability tiers. \textbf{Commercial models} include GPT-5~\cite{openai2025gpt5}, GPT-5 nano~\cite{openai_gpt5_nano_docs_2025}, Claude-Haiku-4.5~\cite{haiku45_2025_news}, GLM-4.5v~\cite{glm45v}, Gemini-2.5-Flash~\cite{google2025gemini25flash}, and Gemini-2.5-Pro~\cite{google2024gemini}. These models represent the cutting edge in general-purpose multimodal reasoning, with strong performance across vision-language tasks. 
\textbf{Open-source models} range from 7B to 72B parameters, include Qwen2.5-VL-7B, Qwen2.5-VL-72B~\cite{Qwen2.5-VL}, Qwen3-VL-8B, Qwen3-VL-30B-A3B~\cite{xu2025qwen3}, InternVL-8B, InternVL-32B~\cite{internvl2024}, LLaVA-v1.5-7B, LLaVA-v1.5-13B~\cite{liu2023llava15}, LLaMA3-LLaVA-Next-8B~\cite{llava_next_interleave_2024}, ShareGPT4V-7B~\cite{chen2024sharegpt4v}. \textbf{Med-specialized models} include MedGemma-27B-IT~\cite{medgemma}, Lingshu-7B~\cite{xu2025lingshu}, LLaVA-Med-7B~\cite{li2025llava-med}, 
HuatuoGPT-Vision-7B, and HuatuoGPT-Vision-34B~\cite{chen2024huatuogpt}. 

\noindent \textbf{Evaluation Metrics.} \quad 
We employ 6 evaluation metrics: \textit{Stage-Wise Accuracy} (\textit{SWA}) reports per-stage correctness; \textit{Chain Completion Rate} (\textit{CCR}) assesses end-to-end reasoning integrity; \textit{Expertise-Stratified Accuracy} (\textit{ESA}) evaluates models across difficulty tiers; \textit{Uncertainty-Aware Score} (\textit{UAS}) rewards calibrated predictions; \textit{Expected Calibration Error} (\textit{ECE}) measures confidence accuracy alignment; and \textit{Modality Contribution Score} (\textit{MCS}) measures each modality's diagnostic importance.


\noindent \textbf{Implementation Details.} \quad 
We adopt unified inference settings: zero-shot prompting with deterministic decoding (temperature=0.0, top\_p=1.0, max\_new\_tokens=256, raised to 16384 for cross-modal tasks). API-hosted models are invoked with these parameters; local models run with fp16 inference on NVIDIA A100 GPUs (single-GPU for \(\leq\)13B and multi-GPU for \(\geq\) 30B models). Confidence calibration (temperature scaling and isotonic regression) is applied only to the validation set, with parameters frozen for testing and assessed via 10-bin ECE. Details are given in Appendix.

\subsection{Progressive Clinical Reasoning}
Table~\ref{tab:main_results} summarizes the evaluation of 21 MLLMs on Progressive Clinical Reasoning, revealing three key findings. 

\noindent \textbf{Commercial Models Lead.} \quad 
Commercial models like GPT-5~\cite{openai2025gpt5}, achieve the highest accuracy (76.24\%), surpassing both open-source models like InternVL-32B~\cite{internvl2024} (70.14\%)) and med-specialized models like MedGemma-27B-IT~\cite{medgemma} (65.21\%). 
In particular, larger models generally exhibit superior reasoning capability (\eg, InternVL-32B~\cite{internvl2024} improves average accuracy from 64.60\% to 70.14\% over InternVL-8B~\cite{internvl2024}). 
All models perform better on R-level than A-level, and A-level than S-level tasks, aligning with expected difficulty progression, \eg, GPT-5~\cite{openai2025gpt5} declines from 80.64\% (R) to 62.92\% (S), while Qwen3-VL-8B~\cite{xu2025qwen3} drops 23.80\%. 
Under uncertainty metrics, Qwen2.5-VL-72B~\cite{Qwen2.5-VL} achieves the highest UAS (66.37\%) among open-source models, yet most exhibit poor calibration, indicating a critical barrier to clinical deployment due to overconfident errors.

\noindent \textbf{Progressive Clinical Reasoning Presents Significant Challenges.} \quad 
Stage-wise accuracy consistently declines across the reasoning chain for all models, due to error propagation and increasing task complexity from image quality assessment to clinical decision-making. For example, GPT-5~\cite{openai2025gpt5} drops from 98.90\% (IQA) to 54.71\% (CD), while the average accuracy across models falls from 80.95\% (QA) to 40.37\% (CD). The most pronounced decrease occurs between DR→SG (–15.32\%), identifying this transition as a critical bottleneck. The Chain Completion Rate (CCR) quantifies a model’s ability to correctly complete all six stages of clinical reasoning in sequence, yet it remains critically low across all evaluated systems. Although CCR values approach zero for some models, even the top-performing GPT-5~\cite{openai2025gpt5} achieves only 24.47\% and fails to complete full reasoning chains in over 75\% of cases, underscoring the substantial challenge of end-to-end clinical reasoning. Notably, Qwen2.5-VL-72B~\cite{Qwen2.5-VL} achieve higher CCR than all commercial models except GPT-5~\cite{openai2025gpt5}, indicating potential for clinical applicability.

\noindent \textbf{MLLMs Lag behind Human Experts.} \quad 
No model surpasses attending physicians (79.91\% overall accuracy), and all remain substantially below specialists (82.85\%), highlighting a significant gap in specialist-level medical reasoning. Although GPT-5~\cite{openai2025gpt5} (76.24\%)  approaches attending-level overall performance, it lags considerably on specialist-level questions (62.92\% vs. 68.27\% for attendings and 83.63\% for specialists). The disparity is most pronounced in chain completion: GPT-5~\cite{openai2025gpt5} achieves 24.47\% CCR versus 62.48\% for specialist, a deficit of 38.52\%. This indicates that while current models handle individual reasoning steps adequately, they struggle with integrated clinical workflows. Specialist' superior performance across difficulty levels (R: 95.88\%, A: 91.28\%, S: 83.63\%) establishes a clear benchmark for future MLLM development.

\begin{table}[t]
\centering
\caption{Evaluation of MLLMs across 6 imaging modalities.} 
\label{tab:fundus_results}
\resizebox{\linewidth}{!}{
\begin{tabular}{clccccccc}
\toprule
& \textbf{MLLM} & \textbf{CFP} & \textbf{OCT} & \textbf{FFA} & \textbf{ICGA} & \textbf{Retcam} & \textbf{EP} & \textbf{Avg} \\
\midrule
\multirow{6}{*}{\rotatebox[origin=c]{90}{\textbf{\textit{Commercial}}}} & 
GPT-5 ~\cite{openai2025gpt5} & 85.42 & 54.81 & 71.47 & 71.47 & 75.28 & 77.12 & 72.60 \\
& Gemini-2.5-Pro ~\cite{google2024gemini} & 80.78 & 51.17 & 71.45 & 71.45 & 65.73 & 69.93 & 68.42 \\
& GLM-4.5v ~\cite{glm45v} & 80.34 & 38.31 & 64.72 & 64.72 & 71.21 & 71.70 & 65.17 \\
& GPT-5-nano~\cite{openai_gpt5_nano_docs_2025} & 84.36 & 49.59 & 60.59 & 60.59 & 64.36 & 69.89 & 64.90 \\
& Gemini-2.5-Flash~\cite{google2025gemini25flash}& 81.91 & 47.95 & 69.97 & 69.97 & 53.20 & 68.89 & 65.31 \\
& Claude-Haiku-4.5 ~\cite{haiku45_2025_news} & 75.28 & 47.45 & 59.80 & 59.80 & 45.76 & 69.27 & 59.56 \\
\midrule
\multirow{10}{*}{\rotatebox[origin=c]{90}{\textbf{\textit{Open-Source}}}} & 
Qwen2.5-VL-72B~\cite{Qwen2.5-VL}& 78.98 & 44.96 & 67.84 & 66.82 & 69.02 & 69.74 & 66.23 \\
& Qwen3-VL-30B-A3B~\cite{xu2025qwen3}& 78.72 & 45.85 & 61.98 & 60.02 & 72.23 & 67.19 & 64.33 \\
& InternVL3-32B ~\cite{internvl2024} & 85.67 & 47.65 & 68.41 & 66.85 & 67.16 & 70.72 & 67.74 \\
& LLaVA-v1.5-13B ~\cite{liu2023llava15} & 48.31 & 27.16 & 38.30 & 39.38 & 65.14 & 45.82 & 44.02 \\
\cmidrule(lr){2-9}
& qwen3-VL-8B~\cite{xu2025qwen3}& 81.18 & 41.62 & 62.99 & 62.59 & 62.79 & 67.45 & 63.10 \\
& Qwen2.5-VL-7B~\cite{Qwen2.5-VL}& 77.58 & 41.62 & 60.99 & 59.94 & 67.22 & 63.47 & 61.80 \\
& LLaMA3-LLaVA-Next-8B ~\cite{llava_next_interleave_2024} & 49.64 & 37.72 & 53.18 & 50.77 & 58.61 & 56.90 & 51.14 \\
& InternVL3-8B ~\cite{internvl2024} & 79.02 & 43.56 & 60.09 & 59.25 & 69.03 & 62.42 & 62.23 \\
& ShareGPT4V-7B~\cite{chen2024sharegpt4v}& 49.27 & 30.82 & 44.29 & 44.32 & 59.01 & 54.61 & 47.05 \\
& LLaVA-v1.5-7B ~\cite{liu2023llava15} & 41.19 & 31.50 & 41.24 & 41.12 & 63.82 & 47.24 & 44.35 \\ 
\midrule
\multirow{5}{*}{\rotatebox[origin=c]{90}{\textbf{\textit{Medical}}}} & 
MedGemma-27B-IT ~\cite{medgemma}& 80.45 & 43.89 & 57.47 & 58.10 & 64.41 & 67.16 & 61.91 \\
& HuaTuoGPT-Vision-34B ~\cite{chen2024huatuogpt} & 70.82 & 36.72 & 53.16 & 53.17 & 65.33 & 52.84 & 55.34 \\
& Lingshu-7B  ~\cite{xu2025lingshu} & 77.24 & 39.20 & 57.65 & 57.20 & 65.51 & 58.98 & 59.30 \\
& LLaVA-Med-7B ~\cite{li2025llava-med}& 61.11 & 21.96 & 31.70 & 31.68 & 35.03 & 38.37 & 36.64 \\
& HuaTuoGPT-Vision-7B ~\cite{chen2024huatuogpt} & 64.13 & 37.34 & 40.78 & 40.79 & 59.77 & 49.49 & 48.72 \\
\bottomrule
\end{tabular}
}
\end{table}

\begin{table}[t]
\centering
\caption{Modality ablation on 58 hospital-collected multi-modal cases. Higher MCS indicates stronger modality dependence. 
} 
\label{tab:modal_contribution}
\resizebox{\columnwidth}{!}{
\begin{tabular}{lcccccc}
\toprule
\multirow{2}{*}{\textbf{Model}} & \multicolumn{5}{c}{\textbf{MCS (Accuracy Drop \%) per Modality}$\downarrow$} & \multirow{2}{*}{\textbf{All.} $\uparrow$}\\
\cmidrule(lr){2-6}
& \textbf{CFP} & \textbf{OCT} & \textbf{FFA} & \textbf{ICGA} & \textbf{EP} & \\
\midrule
\multicolumn{7}{l}{\textit{\textbf{Group 1} Combined CFP/OCT diagnosis. (n=22)}} \\
GPT-5~\cite{openai2025gpt5} & 0.97 & 1.39 & — & — & — & 59.60 \\
Qwen3-VL-30B-A3B~\cite{xu2025qwen3} & 0.28 & 1.20 & — & — & — & 48.07 \\
Qwen3-VL-8B~\cite{xu2025qwen3} & -0.59 & 1.65 & — & — & — & 45.55 \\
MedGemma-27B-IT~\cite{medgemma} & -0.54 & 0.58 & — & — & — & 47.51 \\
\midrule
\multicolumn{7}{l}{\textit{\textbf{Group 2} Combined CFP/OCT/FFA diagnosis. (n=18)}} \\
GPT-5~\cite{openai2025gpt5} & -0.87 & -0.20 & -0.58 & — & — & 59.33 \\
Qwen3-VL-30B-A3B~\cite{xu2025qwen3} & 1.03 & 0.42 & 0.34 & — & — & 49.49 \\
Qwen3-VL-8B~\cite{xu2025qwen3} & -0.42 & -1.01 & -0.25 & — & — & 46.61 \\
MedGemma-27B-IT~\cite{medgemma} & 0.90 & 0.33 & 0.25 & — & — & 46.39 \\
\midrule
\multicolumn{7}{l}{\textit{\textbf{Group 3} Combined CFP/OCT/FFA/ICGA diagnosis. (n=14)}} \\
GPT-5~\cite{openai2025gpt5} & -5.08 & -5.08 & -3.63 & -4.99 & — & 49.37 \\
Qwen3-VL-30B-A3B~\cite{xu2025qwen3} & 0.01 & -1.29 & -0.64 & -4.29 & — & 44.74 \\
Qwen3-VL-8B~\cite{xu2025qwen3} & 0.02 & 0.01 & 0.00 & -3.31 & — & 36.59 \\
MedGemma-27B-IT~\cite{medgemma} & 0.62 & -1.24 & -1.25 & 0.79 & — & 43.33 \\
\midrule
\multicolumn{7}{l}{\textit{\textbf{Group 4} Combined CFP/OCT/FFA/EP diagnosis. (n=4)}} \\
GPT-5~\cite{openai2025gpt5} & -0.56 & -2.38 & -5.76 & — & 2.76 & 47.60 \\
Qwen3-VL-30B-A3B~\cite{xu2025qwen3} & 1.48 & -7.39 & -2.55 & — & 3.79 & 40.70 \\
Qwen3-VL-8B~\cite{xu2025qwen3} & 5.96 & 0.00 & -4.59 & — & 3.23 & 35.41 \\
MedGemma-27B-IT~\cite{medgemma} & 4.97 & 4.06 & 1.02 & — & 2.37 & 41.61 \\
\bottomrule
\end{tabular}
}
\end{table}

\subsection{Cross-Modality Clinical Reasoning}
\noindent \textbf{MLLMs Vary Significantly Across Modalities.} \quad 
Table~\ref{tab:fundus_results} summarizes MLLM performance across imaging modalities. Models perform relatively better on \textit{CFP} and \textit{Retcam}, while accuracy declines significantly on \textit{OCT}, \textit{FFA}, and \textit{ICGA}. 
In particular, OCT/FFA accuracy falls within 20-40\% for smaller open-source and medical-specialized models, indicating that structural (\textit{OCT}) and angiographic (\textit{FFA}/\textit{ICGA}) modalities impose substantially higher demands on fine-grained visual interpretation. 


\noindent \textbf{MLLMs Exhibit Deficiencies in Multi-Modal Clinical Reasoning.} \quad 
We evaluate the cross-modal reasoning capabilities of MLLMs using 58 hospital-sourced clinical cases that require the integration of up to four distinct imaging modalities. Table~\ref{tab:modal_contribution} summarizes the evaluation results for the 58 clinical cases, categorized into four distinct diagnostic modality groups. All evaluated models, including commercial systems such as GPT-5~\cite{openai2025gpt5} and open-source models such as Qwen3-VL~\cite{xu2025qwen3} and LLaVA~\cite{liu2023llava15}, exhibit substantial performance degradation in multi-modal settings relative to single-modality baselines (Table~\ref{tab:fundus_results}). For instance, GPT-5 accuracy declines from 72.60\% to 59.60\%, and Qwen3-VL-30B drops from 64.33\% to 48.07\% in Group 1 cases, underscoring a significant limitation in current MLLMs' ability to effectively integrate multi-modal data for clinical diagnosis. In addition, as diagnostic modalities increase from two (\textit{CFP}/\textit{OCT}) to four (\textit{CFP}/\textit{OCT}/\textit{FFA}/\textit{EP}) across the four groups, model accuracy consistently decreases, indicating that current MLLMs remain inadequate in cross-modal integration and diagnostic reasoning.

\noindent\textbf{Modality Ablation Reveals Ineffective Cross-Modal Integration.} \quad 
We systematically remove each diagnostic modality to assess its impact using the \textit{Modality Contribution Score} (\textit{MCS}), where \textit{MCS}$>$0 indicates performance degradation upon removal and \textit{MCS}$<$0 denotes improvement. Results in Table~\ref{tab:modal_contribution} show heterogeneous MCS patterns across models: while GPT-5~\cite{openai2025gpt5} exhibits moderate performance drops when removing \textit{CFP} or \textit{OCT} in Group 1 cases, Qwen3-VL-8B~\cite{xu2025qwen3} shows accuracy improvement upon \textit{OCT} removal. This inconsistency, coupled with declining overall performance as modality count increases, indicates that current MLLMs struggle to effectively integrate cross-modal information for clinical reasoning.

\section{Conclusion}
X-PCR establishes the first comprehensive benchmark for cross-modality progressive clinical reasoning in ophthalmology. 
Evaluation of 21 leading MLLMs reveals a substantial performance gap in robust clinical reasoning, particularly in progressive reasoning chains and cross-modal diagnostic integration, highlighting the need for significant advancement in clinically-grounded MLLMs. 



\newpage

\section*{Acknowledgement}
This work was supported by the National Natural Science Foundation of China (Grant No. 62576216), Ningbo Municipal Bureau of Science and Technology (Grant Nos. 2023Z138, 2023Z237, 2024Z110, and 2024Z124), and the Guangdong Provincial Key Laboratory (Grant No. 2023B1212060076). The work was also supported by the Intelligent Computing Center of Shenzhen University.

{
    \small
    \bibliographystyle{ieeenat_fullname}
    \bibliography{main}
}


\end{document}